\def\year{2020}\relax
\documentclass[letterpaper]{article} 
\usepackage{aaai20}  
\usepackage{times}  
\usepackage{helvet} 
\usepackage{courier}  
\usepackage[hyphens]{url}  
\usepackage{graphicx} 
\usepackage{graphicx}  
\usepackage{times}
\usepackage{epsfig}
\usepackage{graphicx}
\usepackage{amsmath}
\usepackage{amssymb}

\title{EFANet: Exchangeable Feature Alignment Network for Arbitrary Style Transfer}

\author{Zhijie Wu\textsuperscript{\rm 1}\thanks{Equal contribution. Order determined by coin toss.}, Chunjin Song\textsuperscript{\rm 1}$^*$, Yang Zhou\textsuperscript{\rm 1}$^{\dagger}$, Minglun Gong\textsuperscript{\rm 2}, Hui Huang\textsuperscript{\rm 1}\thanks{Corresponding authors.}\\
\textsuperscript{\rm 1} Shenzhen University, \qquad \textsuperscript{\rm 2} University of Guelph\\
\{wzj.micker, songchunjin1990, zhouyangvcc\}@gmail.com, \\minglun@uoguelph.ca,\quad hhzhiyan@gmail.com\\}

\usepackage[dvipsnames]{xcolor}

\definecolor{turquoise}{cmyk}{0.65,0,0.1,0.1}
\definecolor{purple}{rgb}{0.65,0,0.65}
\definecolor{dark_green}{rgb}{0, 0.5, 0}
\definecolor{orange}{rgb}{0.8, 0.6, 0.2}
\definecolor{red}{rgb}{0.8, 0.2, 0.2}
\definecolor{blue}{rgb}{0, 0, 1}
\definecolor{brown}{rgb}{0.5, 0.16, 0.16}

\newcommand{\zj}[1]{#1}

\begin{document}

\maketitle

\begin{abstract}
	Style transfer has been an important topic both in computer vision and graphics. Since the seminal work of Gatys et al. first demonstrates the power of stylization through optimization in the deep feature space, quite a few approaches have achieved real-time arbitrary style transfer with straightforward statistic matching techniques. In this work, our key observation is that only considering features in the input style image for the global deep feature statistic matching or local patch swap may not always ensure a satisfactory style transfer; see e.g., Figure~\ref{Fig:teaser_comparison}. Instead, we propose a novel transfer framework, EFANet, that aims to jointly analyze and better align exchangeable features extracted from content and style image pair.
	In this way, the style features from the style image seek for the best compatibility with the content information in the content image, leading to more structured stylization results. 
	In addition, a new whitening loss is developed for purifying the computed content features and better fusion with styles in feature space.
	Qualitative and quantitative experiments demonstrate the advantages of our approach.
\end{abstract}

\section{Introduction}

A style transfer method takes a pair of images as input and synthesize an output image that preserves the content of the first image while mimicking the style of the second image. The study on this topic has drawn much attention in recent years due to its scientific and artistic values. Recently, the seminal work~\cite{gatys2016image} found that multi-level feature statistics extracted from a pre-trained CNN model can be used to separate content and style information, making it possible to combine content and style of arbitrary images. This method, however, depends on a slow iterative optimization, which limits its range of application.

Since then, many attempts have been made to accelerate the above approach through replacing the optimization process with a feed-forward neural networks~\cite{Dumoulin2016ALR,johnson2016perceptual,li2017diversified,Ulyanov2016TextureNF,zhang2017MultistyleGN}.
While these methods can effectively speed up the stylization process, they are generally constrained to a predefined set of styles and cannot adapt to an arbitrary style specified by a single exemplar image.

\begin{figure}[t]
	\centering
	\includegraphics[width=\linewidth]{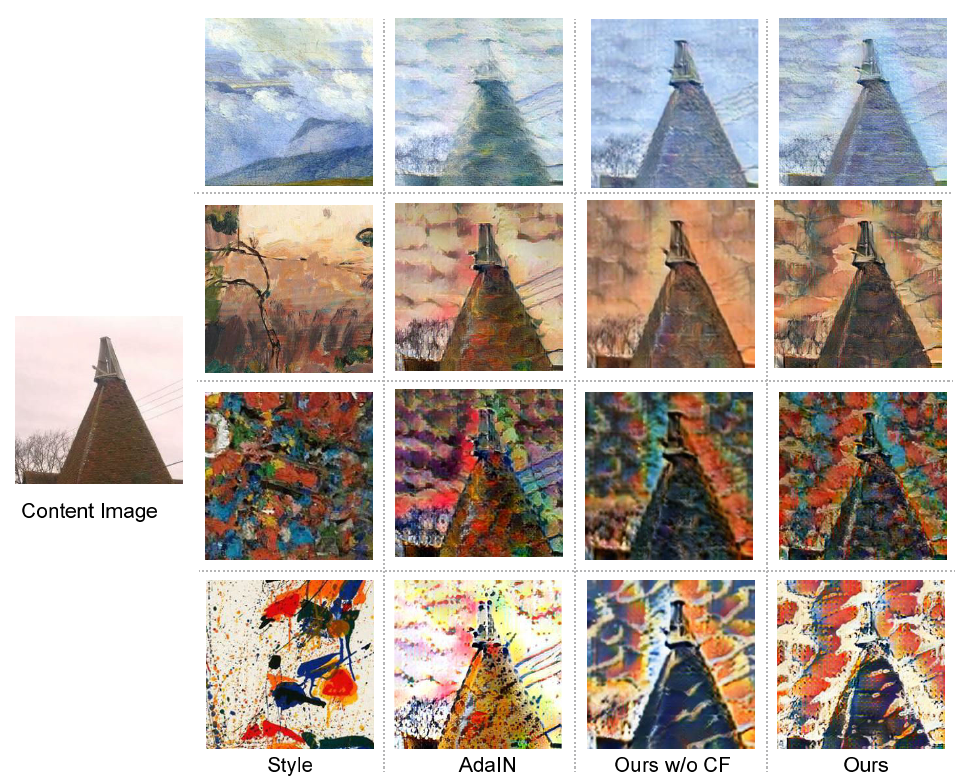}
	\caption{
	The existing method (AdaIN) ignores differences in style images while our approach jointly analyzes each content-style image pair and computes exchangeable style features. As a result, AdaIN and the baseline model without common features (4th column) only work well with a simple style (1st and 2nd row). 
	When the target styles become more complex and the content-style images have different patterns/color distributions, AdaIN and the baseline model fail to capture the salient style patterns and suffer from insufficiently stylized results (color distribution and textures in 3rd \& 4th row).  
	In comparison, our model better adapts to pattern/color variation in the content image and map compatible patterns/colors in the style images accordingly.}
	\label{Fig:teaser_comparison}
\end{figure}

Notable efforts~\cite{Chen2016FastPS,Huang2017ArbitraryST,Li2017UniversalST,Shen2017MetaNF,Sheng2018AvatarNetMZ} have been devoted to solving this flexibility v.s. speed dilemma. A successful direction is to apply statistical transformation, which aligns feature statistics of the input content image to that of the style image~\cite{Huang2017ArbitraryST,Li2017UniversalST,Sheng2018AvatarNetMZ}. 
However, as shown in Figure~\ref{Fig:teaser_comparison}, the style images can be dramatically different from each other and from the content image, both in terms of semantic structures and style features.
Performing style transfer through statistically matching different content images to the same set of features extracted from the style image often introduces unexpected or distorted patterns~\cite{Huang2017ArbitraryST,Li2017UniversalST}.
Several methods~\cite{Sheng2018AvatarNetMZ,yao2019attention,Park_2019_CVPR} conquer these disadvantages through patch swap with a multi-scale feature fusion, but may contain spatially distorting semantic structures when the local patterns from input images differ a lot.

To address the aforementioned problems, in this paper, 
we jointly consider both content and style images and extract \emph{common} style features, which is customized for this pair of images only. 
Through maximizing the common features, our goal is to align the style features of both content and style images as much as possible. This follows the intuition that when the target style features are 
compatible with the content image, we can get good transfer result. Since the style features of content image are computed from its own content information, they are definitely compatible with each other.  Hence aligning the style features of the two images helps to improve the final stylization; see the comparison of our method with \& without common feature in Figure~\ref{Fig:teaser_comparison}. 

Intuitively, the common style features we extracted bridge the gap between the input content and style images, making our method outperform existing methods in many challenging scenarios. We call the aligned style features as \emph{exchangeable} style features.
Experiments demonstrate that performing style transfer based on our exchangeable style features yields more structured results with better visual style patterns than existing approaches; see e.g., Figures~\ref{Fig:teaser_comparison} and~\ref{Fig:comparison}.

To compute exchangeable style features from feature statistics of two input images, a novel \emph{Feature Exchange Block} is designed, which is inspired by the works on private-shared component analysis~\cite{Bousmalis2016DomainSN,cao2018dida}. In addition, we propose a new \emph{whitening loss} to facilitate the combination between content and style features by removing style patterns existed in content images. To summarize, the contributions of our work include:

\begin{itemize}
	\item{The importance of aligning style features} for style transfer between two images is clearly demonstrated.
	\item{A novel Feature Exchange Block as well as a constraint loss function are designed for the pair-wise analysis of learning common information in-between style features.}
	\item{A simple yet effective \emph{whitening loss} is developed to encourage the fusion between content and style information by filtering style patterns in content images.}
	\item{The overall end-to-end style transfer framework can perform arbitrary style transfer in real-time and synthesize high-quality results with favored styles.}
\end{itemize}

\section{Related Work}

\paragraph{Fast Abitrary Style Transfer}

Intuitively, style transfer aims at changing the style of an image while preserving its content. Recently, impressive style transfer is realized by Gatys et al.~\citeyear{gatys2016image} based on deep neural networks. Since then, many methods are proposed to train a single model that can transfer arbitrary styles. Here we only review the related works on arbitrary style transfer and refer the readers to~\cite{jing2017neural} for a comprehensive survey.

Chen et al.~\citeyear{Chen2016FastPS} realize the first fast neural method by matching and swapping local patches between the intermediate features of content and style images, which is thus called Style-Swap. Then Huang et al.~\citeyear{Huang2017ArbitraryST} propose an adaptive instance normalization (AdaIN) to explicitly match the mean and variance of each feature channel of the content image to those of the style image. Li et al.~\citeyear{Li2017UniversalST} further apply whitening and coloring transform (WCT) to align the correlations between the extracted deep features. Sheng et al.~\citeyear{Sheng2018AvatarNetMZ} develop Avatar-Net to combine local and holistic style pattern transformation, achieving better stylization regardless of the domain gap. Very recently, AAMS (Yao et al.~\citeyear{yao2019attention}) tries to transfer the multi-stroke patterns by introducing self-attention mechanism. Meanwhile, SANet~\cite{Park_2019_CVPR} promotes Avatar-Net by learning a similarity matrix and flexibly matching the semantically nearest style features onto the content features. And Li et al.~\citeyear{Li_2019_CVPR} speeds up WCT with a linear propagation module. \zj{In order to boost the generalization ability, ETNet~\cite{song2019etnet} evaluate errors in the synthesized results and correct them iteratively.}
The above methods, however, all achieve stylization by a straightforward statistic matching or local patch matching and ignore the gaps between input features, which may not be able to adapt to the unlimited variety. 

In this paper, we still follow the holistic alignment with respect to feature correlations. The key difference is that before applying style features, we jointly analyze the similarities between the style features of content and style images. Thus these style features can be aligned accordingly, which enables the style features to match the content images more flexibly and improves the final compatibility level between target content and style features significantly.

\paragraph{Feature Disentanglement}
Learning disentangled representation aims at separating the learned internal representation into the factors of data variations~\cite{whitney2016disentangled}. It improves the re-usability and interpretation of the model, which is very useful for e.g., domain adaptation~\cite{Bousmalis2016DomainSN,cao2018dida}. 
Recently, based on the works on generative models~\cite{goodfellow2014generative,cao18llc}, several concurrent works~\cite{Lee2018DiverseIT,Huang2018MultimodalUI,Gonzalez2018NIPS,Ma2018ExemplarGuided,cao2019multi} have been proposed for multi-modal image-to-image translation. They map the input images into one common feature space for content representation and two unique feature spaces for styles.  Yi et al.~\citeyear{yi2018branched} design BranchGAN to achieve scale-disentanglement in image generation.
Wu et al.~\citeyear{SAGnet19} advance 3D shape generation by disentangling geometry and structure information.
For style transfer, some efforts~\cite{zhang2018separating,zhang2018style} are also made to separate a representation of one image into the content and style. Different from the mentioned methods, we perform feature disentanglement only on style features of the input image pair. A common component is thus extracted, which is then used to compute exchangeable style features for style transfer.

\section{Developed Framework}

\begin{figure}[t]
	\centering
	\includegraphics[width=\linewidth]{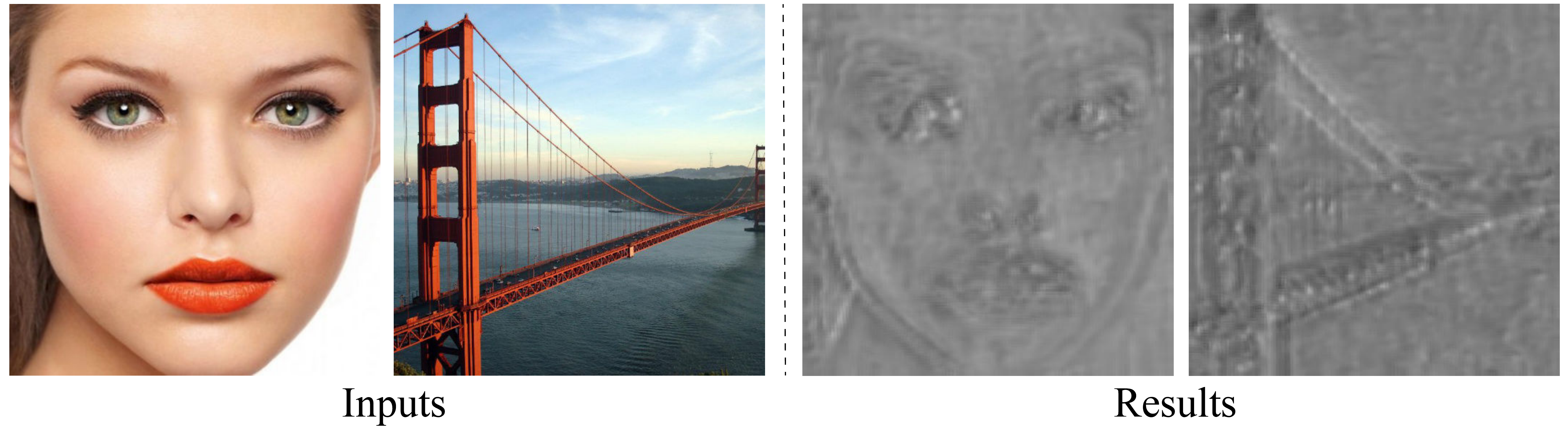}
	\caption{Images decoded from whitened features. The results on the right are rescaled for better visualization. The whitened features still keep spatial structures but various style patterns are removed.}
	\label{Fig:purify_content}
\end{figure}

Following~\cite{gatys2016image}, we consider the deep feature extracted by the network pretrained on large dataset as the content representation for an image, and the feature correlation at a given layer as the style information. By fusing the content feature with a new target style feature, we can generate a stylized image. 

The overall goal of our framework is to better align style features between the style and content images, such that the style features from one image can better match the content of the other image,  resulting a better stylization adaptively. To achieve that, a key module of Feature Exchange block  is proposed to jointly analyze the style features of the two input images. A common feature is disentangled to encode the shared components between the style features, \zj{indicating the similarity information among them. Then with the common features as guiders, we can make the target style features be more similar to the input contents and facilitate the alignment between them.}

\begin{figure*}[t!]
	\centering
	\includegraphics[width=\linewidth]{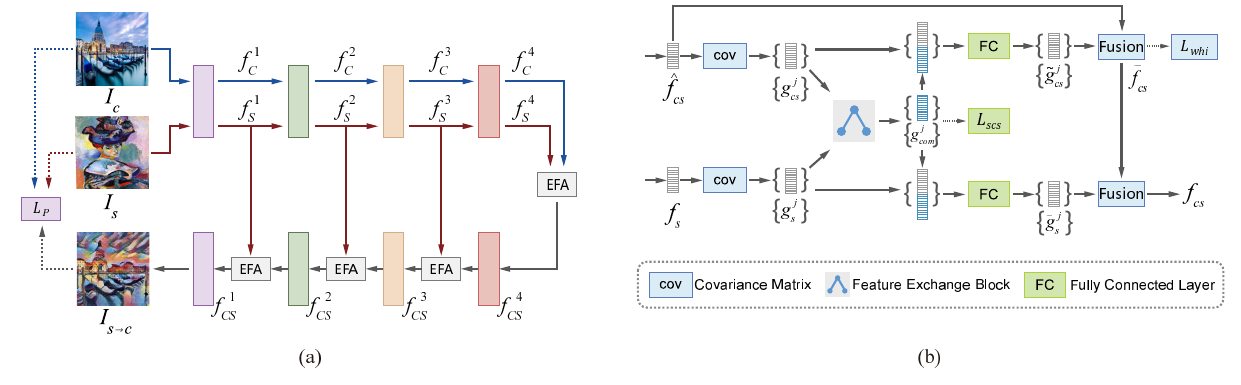}
	\caption{(a) Architecture overview. The input image pair $I_c$ and $I_s$, goes through the pre-trained VGG encoder to extract feature maps $\{f^i_c\}$ and $\{f^i_s\}, i\in\{1,...,L\},L=4$. Then, starting from $f^L_c$ and $f^L_s$, different EFANet modules are applied to progressively fuse styles into corresponding decoded features for final stylized images.
	(b) The architecture of EFANet module. Given $\hat{f}_{cs}$ and $f_s$ as inputs, we compute two Gram matrices as the raw styles and then represent them as two lists of feature vectors $\{g_{cs}^j\}$ and $\{g_s^j\}$. Each corresponding style vector pair ($g_{cs}^j$ and $g_s^j$) is fed into the newly proposed Feature Exchange Block and a common feature vector $g_{com}^j$ is extracted via the joint analysis. We concatenate $g_{com}^j$ with $g_{cs}^j$ and $g_s^j$ respectively to learn two exchangeable style feature $\tilde{g}_{cs}^j$ and $\tilde{g}_s^j$.
	$\tilde{g}_{cs}^j$ is used for the content feature purification, which will be further fused with  $\tilde{g}_s^j$, outputting $f_{cs}$. Finally $f_{cs}$ will be either propagated for finer-scale information or decoded into stylized images $I_{s\rightarrow c}$.}
	\label{Fig:architecture}
\end{figure*} 

\begin{figure}[t!]
	\centering
	\includegraphics[width=\linewidth]{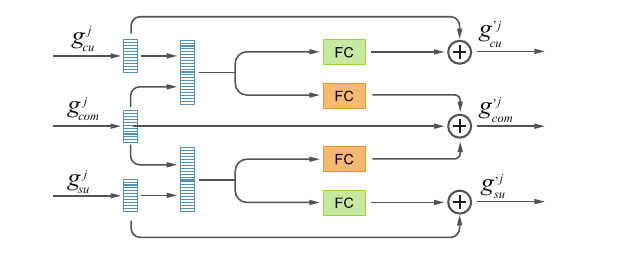}
	\caption{Architecture of a Feature Exchange Block, where $\oplus$ denotes element-wise addition. Each block has three input features, one common feature $g^j_{com}$ and two unique features for content $g^j_{cu}$ and style $g^j_{su}$ images, respectively. 
	\zj{The features, $g_{cu}^j$ and $g_{su}^j$, are first initialized with $g_{cs}^j$ and $g_s^j$ respectively and the $g_{com}^j$ with their combination.}
	Then the block allows common feature to interact with unique features and outputs refined results ${g'}^j_{com}$, ${g'}^j_{cu}$, and ${g'}^j_{su}$.}
	\label{Fig:proposed_block}
\end{figure}

\subsection{Exchangeable Feature for Style Transfer}
\label{subsec:framework}

As illustrated in Figure~\ref{Fig:architecture}(a), our framework mainly consists of three parts: one encoder, several EFANet modules ($\Omega(\cdot)$) and one decoder for generating the final images. We denote $f^i_c$ and $f^i_s$, $i \in \{1,...,L\}, L=4$ as the feature maps outputted by the $relu\_i$ layer of the pre-trained VGG encoder, which correspond to content and style images ($I_c$ and $I_s$) respectively. We equip the multi-scale style adaption strategy to advance the stylization performance. Specifically, in the bottleneck of the conventional encoder-decoder architecture, starting from $f^L_c$ and $f^L_s$, different EFANet modules are applied to progressively fuse the styles from input images into the corresponding decoded features in a coarse-to-fine manner as $f^i_{cs} = \Omega(\hat{f}^{i}_{cs}, f^i_s)$. The $f^i_{cs}$ indicates a decoded stylized feature and $\hat{f}^{i}_{cs} = u(f^{i+1}_{cs})$, where $u(\cdot)$ is an upsampling operator and the superscript $i$ denotes the $i$-th scale. Note that, initially we set $\hat{f}^{L}_{cs} = f^L_c$ and apply the superscript $j$ to indicate the $j$-th style vector of a Gram matrix in the following paragraphs.

In Figure~\ref{Fig:architecture}(b), given $f_s$ and $\hat{f}_{cs}$ as inputs,
we first compute two Gram matrices across the feature channels as the raw style representations and denote them as $G_s$ and $G_{cs} \in R^{C \times C}$. The $C$ indicates the channel number for $f_s$ and $\hat{f}_{cs}$.
In order to preserve more style details in output results and reduce computation burden, we process only a part of style information at a time and represent $G_s$ and $G_{cs}$ as two lists of style vectors, e.g. $G_s=\{g_s^{1},g_s^{2},...,g_s^{C}\}$ and $G_{cs}=\{g_{cs}^{1},g_{cs}^{2},...,g_{cs}^{C}\}$.
Each style vector, $g_{s}^{j}$ and $g_{cs}^{j}$, compactly encodes the mutual relationships between the $j$-th channel and the whole feature map.
Then each corresponding style vector pair ($g_{s}^{j}$, $g_{cs}^{j}$) is processed using one Feature Exchange block.
Accordingly a common feature $g_{com}^j$ and two unique feature vectors for decoded information (as content) and style, $g_{cu}^j$ and $g_{su}^j$, can be disentangled. 

Guided by $g_{com}^j$, the style features are aligned in the following manner: we first concatenate $g_{com}^j$ with the raw style vectors $g_s^j$ and $g_{cs}^j$ respectively. Then they are sent into fully connected layers individually, yielding the aligned style vectors $\tilde{g}_s^j$ and $\tilde{g}_{cs}^j$. We call them as exchangeable style features since each of them can be used easily to adapt its style to the target image. Then we stack the style vectors $\{\tilde{g}_s^j\}$ and $\{\tilde{g}_{cs}^j\}$ into two matrices, $\tilde{G}_s$ and $\tilde{G}_{cs}$, for later fusion as:
\begin{eqnarray}
\nonumber
\tilde{G}_s = [\tilde{g}_s^1, \tilde{g}_s^2,...,\tilde{g}_s^C], \quad
\tilde{G}_{cs} = [\tilde{g}_{cs}^1, \tilde{g}_{cs}^2,...,\tilde{g}_{cs}^C].
\end{eqnarray}

Inspired by the whitening operation of WCT~\cite{Li2017UniversalST}, we assume that better stylization results can be achieved when the target content features are uncorralated before content-style fusion. The whitening operation can be regarded as a function, where the content feature is filtered by its corresponding style info.
Thus after the feature alignment, to facilitate transferring a new style to an image, we use the exchangeable style to purify its own content feature through a fusion as:
\begin{equation}
\nonumber
\tilde{f}_{cs} = \Psi_{whi}(\hat{f}_{cs}, \tilde{G}_{cs}) = \hat{f}_{cs} \cdot W_{whi} \cdot \tilde{G}_{cs},
\end{equation}
where $\Psi_{whi}(\cdot)$ and $W_{whi}$ indicates the fusion operation and a learnable matrix respectively~\cite{zhang2018separating,zhang2017MultistyleGN}. 
Moreover, we develop a \emph{whitening loss} to further encourage the removal of correlations between different channels;
see Figure~\ref{Fig:purify_content} as a validating example. The details of the whitening loss are discussed in the Loss Function section below.

Finally, we exchange the aligned style vectors and fuse them with the purified content features as:
\begin{equation}
\nonumber
f_{cs} = \Psi_{fusion}(\tilde{f}_{cs}, \tilde{G}_{s}) = \tilde{f}_{cs} \cdot W_{fusion} \cdot \tilde{G}_{s}.
\end{equation}
Then the $f_{cs}$ will be propagated to receive style information at finer scales or decoded to output stylized images. The decoder is trained to learn the inversion from the fused feature map to image space, and hereby, style transfer is eventually achieved for both input images. Note that the resulting $I_{s\rightarrow c}$ denotes the stylization image that transfers style in $I_s$ to $I_c$,

\subsection{Feature Exchange Block}
\label{subsec:block}

According to Bousmalis et al.~\shortcite{Bousmalis2016DomainSN}, explicitly modeling the unique information would help improve the extraction of the shared component. To adapt this idea for our exchangeable style features, a Feature Exchange block is proposed to jointly analyze the style features of both input images and model their inter-relationships, based on which we explicitly update the common feature and two unique features for the disentanglement.
Figure~\ref{Fig:proposed_block} illustrates the detailed architecture, where the unique features, $g_{cu}^j$ and $g_{su}^j$, are first initialized with $g_{cs}^j$ and $g_s^j$ respectively and the $g_{com}^j$ with their combination. 
Then they are updated by the learned residual features. Using residual learning is to facilitate gradient propagation during training and convey messages so that each input feature can be directly updated. This property allows us to chain any number of Feature Exchange blocks in a model, without breaking its initial behavior.

As shown in Figure~\ref{Fig:proposed_block}, there are two shared fully-connected layers inside each block. To be specific, the disentangled features are updated as:
\begin{eqnarray}
\nonumber
	{g'}_{com}^j &=& \Theta_{com}([g_{cu}^j, g_{com}^j]) + \Theta_{com}([g_{su}^j, g_{com}^j]) + g_{com}^j,\\ \nonumber
	{g'}^{j}_{cu} &=& \Theta_{uni}([g_{cu}^j, g_{com}^j]) + g_{cu}^j,
\end{eqnarray}
where $\Theta_{com}(\cdot)$ and $\Theta_{uni}(\cdot)$ denote the fully-connected layers to output residuals for the common features and unique features respectively. $[\cdot, \cdot]$ indicates a concatenation operation. We can update ${g}_{su}^{j}$ in a similar way.

By doing so, the feature exchange blocks enable $g_{com}^j$ and $g_{cu}^j$ (or $g_{su}^j$) to interact with each other by modelling their dependencies and thus to be refined to the optimal.

On the other hand, to make sure the feature exchange block conduct proper disentanglement, a constraint on the disentangled feature is added following Bousmalis et al.~\shortcite{Bousmalis2016DomainSN}. First, $g_{com}^j$ should be orthogonal to both $g_{cu}^j$ and $g_{su}^j$ as much as possible. Meanwhile, it should let us be able to reconstruct $g_s^j$ and $g_{cs}^j$ based on the finally disentangled features. Therefore, a feature exchange loss can be defined as:
\begin{equation*}
L_{ex}^j = g_{com}^j\cdot g_{cu}^j + g_{com}^j\cdot g_{su}^j + \|g_{cs}^j - \bar{g}_{cs}^j \|_{1} + \|g_s^j - \bar{g}_{s}^j\|_{1},
\end{equation*}
where $\bar{g}_{cs}^j$ is the reconstructed style vector by feeding the sum of $g_{com}^j$ and $g_{cu}^j$ into a fully connected layer. $\bar{g}_{s}^j$ is the reconstruction from $g_{com}^j$ and $g_{su}^j$. Note that this fully connected layer for reconstruction is only valid in training stage, and $L_{ex}^j$ is only computed with the final output of the feature exchange block. And we use only one feature exchange block in each EFANet module.

Finally, to maximize the common information, we also penalize the amount of unique features. Thus the final loss function for the common feature extraction is:
\begin{equation*}
\label{eq:L_com}
L_{com} = \sum^{C}_{j=1} L_{ex}^j + \lambda^{uni} (\|g_{cu}^j\|^2 + \|g_{su}^j\|^2),
\end{equation*}
where $\|\cdot\|$ denotes $L_2$ norm of a vector, and $\lambda^{uni}$ is set to 0.0001 in all our experiments.

\subsection{Loss Function for Training}
As illustrated in Figure~\ref{Fig:architecture}, three different types of losses are computed for each input image pair. 
The first one is perceptual loss~\cite{johnson2016perceptual}, which is used to evaluate the stylized results. Following previous work~\cite{Huang2017ArbitraryST,Sheng2018AvatarNetMZ}, we employ a VGG model~\cite{Simonyan2014VeryDC} pre-trained on ImageNet~\cite{deng2009imagenet} to compute the perceptual content loss:
\begin{equation*}
\begin{aligned}
L^{c}_{p} = \left \|E(I_{c})-E(I_{s\rightarrow c})\right \|_{2},
\end{aligned}
\end{equation*}
and style loss:
\begin{equation*}
\begin{aligned}
L^{s}_{p} = \sum^{L}_{i=1}{\left \|G^{i}(I_{s}) - G^{i}(I_{s\rightarrow c})\right \|_{2}},
\end{aligned}
\end{equation*}
where $E(\cdot)$ denotes the VGG-based encoder and $G^{i}(\cdot)$ represents a Gram matrix for features extracted at $i$-th scale in the encoder module. As mentioned before, we set $L=4$.

The second is the \emph{whitening loss}, which is used to remove style information in target content images at training stages. According to Li et al.~\shortcite{Li2017UniversalST}, after the whitening operation, $\tilde{f}_{cs} \cdot (\tilde{f}_{cs})^\mathrm{T}$ should equal the identity matrix. Thus we define the \emph{whitening loss} as:
\begin{equation*}
L_{whi} = \|\tilde{f}_{cs} \cdot (\tilde{f}_{cs})^\mathrm{T} - I\|_2
\end{equation*}
where $I$ denotes the identity matrix. By doing so, we can encourage feature map $\tilde{f}_{cs}$ to be as uncorrelated as possible.

The third one is the common feature loss, $L_{com}$, defined previously for a better feature disentanglement.

Note that, for both $L_{whi}$ and $L_{com}$, we sum up the losses over all scales, e.g. $L_{whi} = \sum^L_{i=1} L^i_{whi}$ and $L_{com} = \sum^L_{i=1} L^i_{com}$. The superscript $i$ here indicates losses computed at $i$-th scale, where $i \in \{1,..,L\}$.  To summarize, the full objective function of our proposed network is:
\begin{equation}
\begin{aligned}
L_{total} = \lambda^{pc}L^{c}_{p} + \lambda^{ps}L^{s}_{p} + \lambda^{whi} L_{whi}+\lambda^{com} L_{com}, \nonumber
\end{aligned}
\label{subsec:loss_function}
\end{equation}
where the four weighting parameters are respectively set as 1, 7, 0.1 and 5 through out the experiments.

\subsection{Implementation Details}

We implement our model with Tensorflow~\cite{abadi2016tensorflow}. In general, our framework consists of an encoder, several EFANet modules and a decoder. Similar to prior work~\cite{Huang2017ArbitraryST,Sheng2018AvatarNetMZ}, we use the VGG-19 model~\cite{Simonyan2014VeryDC} (up to relu4\_1) pre-trained on ImageNet~\cite{deng2009imagenet} to initialize the fixed encoder. For the decoder, after the fusion of style and content features, two residual blocks are used, followed by upsampling operations. Nearest-neighbor upscaling plus convolution strategy is used to reduce artifacts in the upsampling stage~\cite{odena2016deconvolution}. We choose Adam optimizer~\cite{Kingma2014AdamAM} with a batch size of 4 and a learning rate of 0.0001, and set the decay rates by default for 150000 iterations.

Place365 database~\cite{Zhou2014LearningDF} and WiKiArt dataset~\cite{nich2016wikiart} are used for content and style images respectively, following~\cite{Sanakoyeu2018ASC}. During training, we resize the smaller dimension of each image to 512 pixels with the original image ratio. Then we train our model with randomly sampled patches of size $256 \times 256$. Note that in the testing stage, both the content and style images can be of any size.

\label{subsec:training}
\section{Experimental Results}
\label{Sec:experiments}

\begin{figure*}[t]
	\centering
	\includegraphics[width=\linewidth]{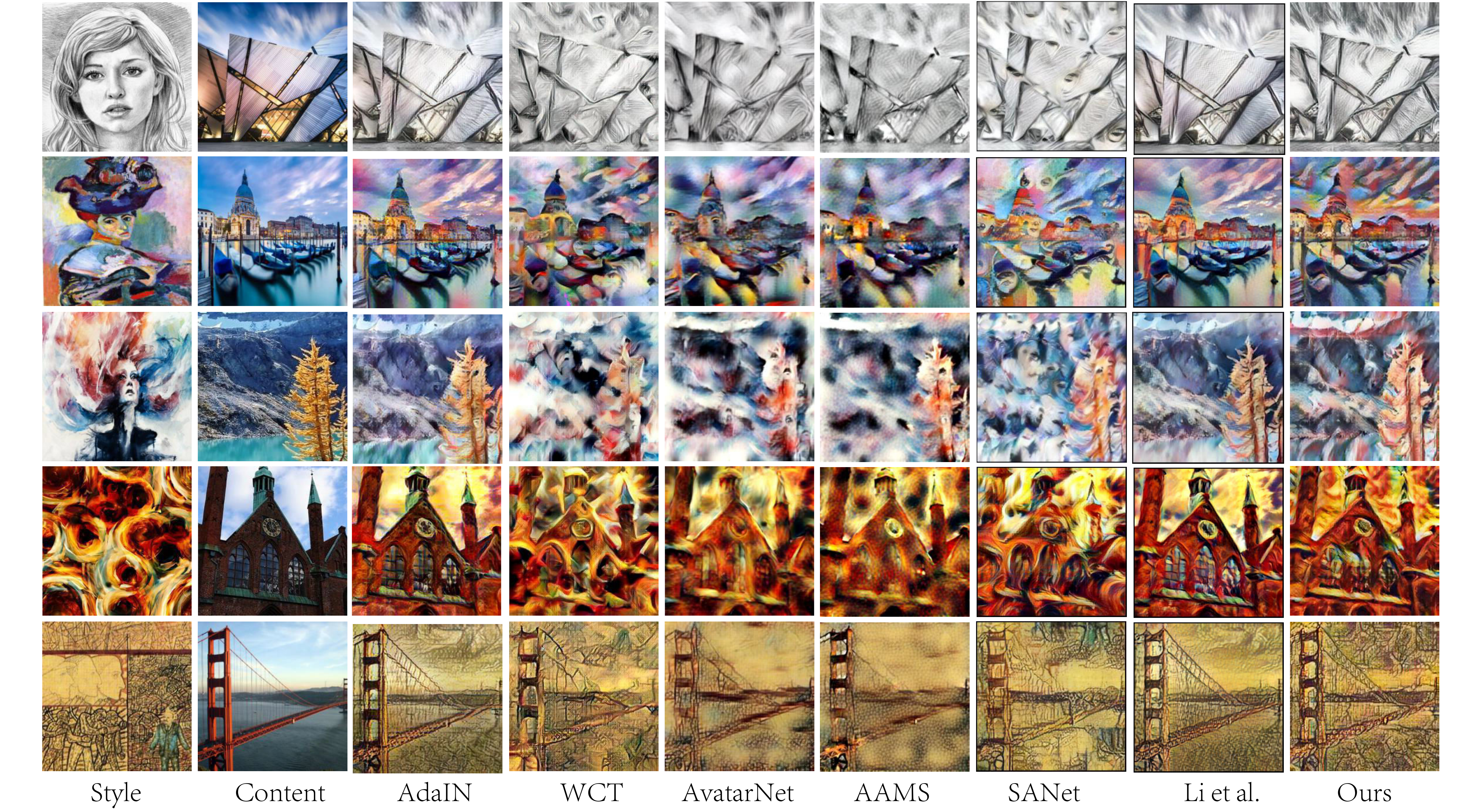}
	\caption{Comparison with results from different methods. Note that the proposed model generates images with better visual quality while the results of other baselines have various artifacts; see text for detailed discussions.}
	\label{Fig:comparison}
\end{figure*} 

\begin{table*}[]
	\caption{Quantitative comparison over different models on perceptual (content \& style) loss, preference score of user study and running time. Note that all the results are averaged over 100 test images except the preference score. The $Ours^{*}$ denotes a model equiped with single-scale strategy.}
	\small
	\label{tab:p_loss}
	\begin{center}
		\begin{tabular}{cccccccccc}
			\hline
			Loss     & AdaIN & WCT & Avatar-Net & AAMS & SANet & Li et al. &  Ours w/o CF & Ours$^{*}$ & Ours \\ \hline
			Content ($L_c$)   & 14.4226  & 19.5318  & 16.8482 & 17.1321 & 23.3074 & 18.7288 & 16.3763 & 16.8600 & 16.5927  \\ 
			Style ($L_s$)     & 40.5989  & 27.1998  & 31.1532 & 34.7786 & 29.7760 & 37.3573 & 22.6713 & 24.9123 & 14.8582  \\ 
			Preference/\%     & 0.110  & 0.155  & 0.150 & 0.137 & 0.140 & 0.108 & - & - & 0.200  \\ 
			Time/sec     & 0.0192  & 0.4268  & 0.9258 & 1.1938 & 0.0983 & 0.0071 & 0.0227 & 0.0208 & 0.0234  \\ \hline
		\end{tabular}
	\end{center}
	\vspace{-0.15in}
\end{table*}

\paragraph{Comparison with Existing Methods}

We compare our approach with six state-of-the-art methods for arbitrary style transfer: AdaIn~\cite{Huang2017ArbitraryST}, WCT~\cite{Li2017UniversalST}, Avatar-Net~\cite{Sheng2018AvatarNetMZ}, AAMS~\cite{yao2019attention}, SANet~\cite{Park_2019_CVPR} and Li et al.~\cite{Li_2019_CVPR}. 
For the compared methods, publicly available codes with default configurations are used for a fair comparison.

Results of qualitative comparisons are shown in Figure~\ref{Fig:comparison}. 
For the holistic statistic matching pipelines, AdaIN~\cite{Huang2017ArbitraryST} can achieve arbitrary style transfer in real-time.
However, it does not respect semantic information and sometimes generates less stylized results with color distribution different from the style image (see row 1 \& 3). WCT~\cite{Li2017UniversalST} improves the stylization a lot but often introduces distorted patterns. As shown in rows 3 \& 4, it sometimes produces messy and less-structured images.  
Li et al.~\citeyear{Li_2019_CVPR} proposes a linear propagation module and achieves the fastest transfer among all the compared methods. But it often gets stuck into the instylization issuses and can not adapt the compatible style patterns or color variations to results (row 1 \& 3).

Then Avatar-Net~\cite{Sheng2018AvatarNetMZ} improves over the holistic matching methods by adapting more style details to results with a feature decorating module, but it also blurs the semantic structures (rows 3) and sometimes distorts the salient style patterns (see rows 1 \& 5).
While AAMS~\cite{yao2019attention} stylizes images with multi-stroke style patterns, similar to Avatar-Net, it still suffers from the structure distortion issues (row 3) and introduces unseen dot-wise artifacts (row 2 \& 5). It also fails to capture the patterns presented in style image (row 5).  In order to match the semantically nearest style features onto the content features, SANet~\cite{Park_2019_CVPR} shares the similar spirits with Avatar-Net but employs a style attention module in a more flexible way. Thus it might still blur the content structures (row 3) and directly copy some semantic patterns in content images to stylization results (e.g. the eyes in row 1, 2 \& 3). Due to the local patch matching, SANet also distorts the presented style patterns and fails to reserve the texture consistency (row 5).

In contrast, our approach achieves more favorable performance. The alignment on style features allows our model to better match the regions in content images with patterns in style images. The target style textures can be adaptively transferred to the content images, manifesting superior texture detail (last row) and richer color variation (2nd row). Compared to most methods, our approach can also generate more structured results  while the style pattern, like brush strokes, is preserved well (3rd row).

\begin{figure*}[t]
	\centering
	\includegraphics[width=\linewidth]{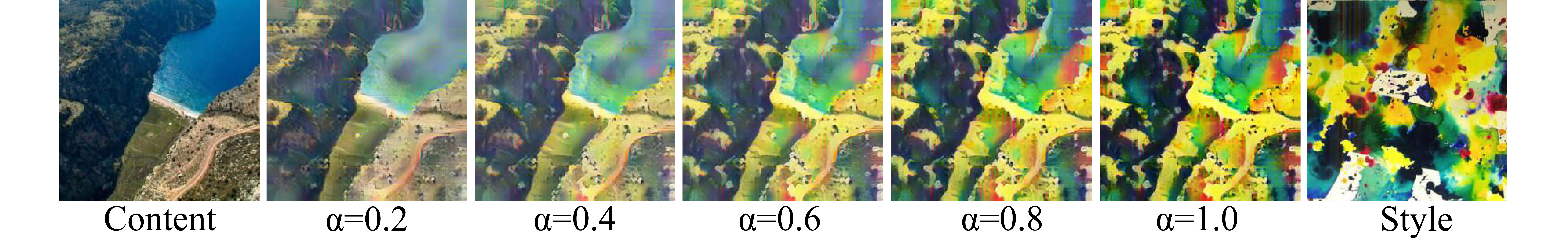}
	\caption{Balance between content and style. At testing stage, the degree of stylization can be controlled using parameter $\alpha$.}
	\label{Fig:application_tradeOff}
\end{figure*}

\begin{figure}[t]
	\centering
	\includegraphics[width=\linewidth]{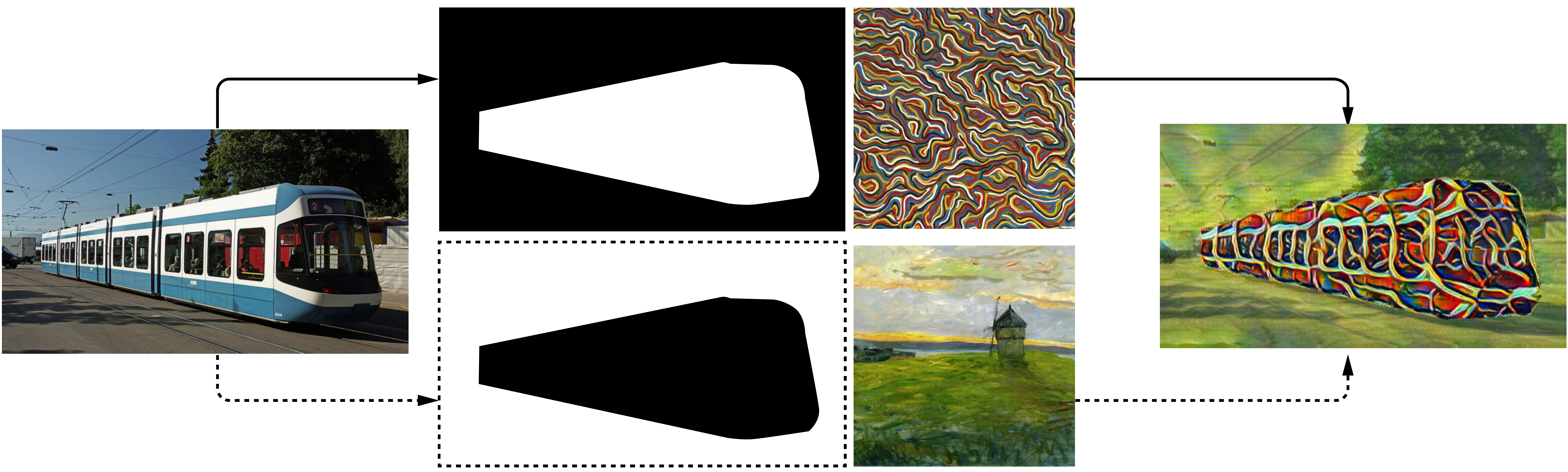}
	\caption{Application for spatial control. Left: content image. Middle: style images with masks to indicate target regions. Right: synthesized result.}
	\label{Fig:application_mask}
\end{figure}

\begin{figure}[t]
	\centering
	\includegraphics[width=\linewidth]{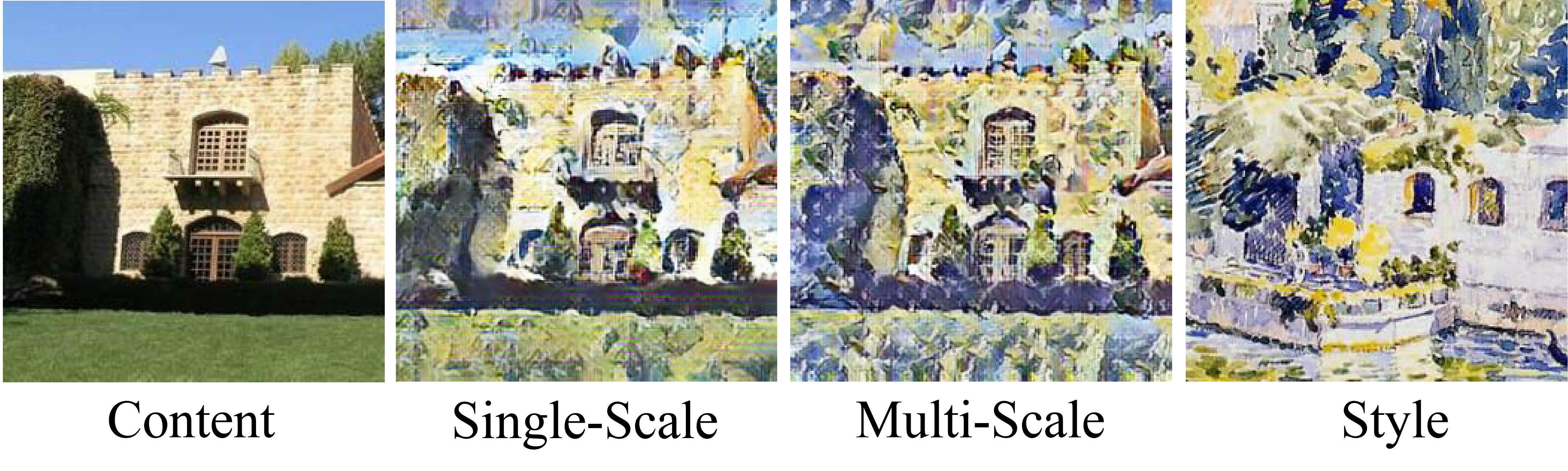}
	\caption{Ablation study on multi-scale strategy. By fusing the content and style in multi-scales, we can enrich the local and global style patterns for stylized images.}
	\label{Fig:ab_multi_scale}
\end{figure}

\begin{figure}[t]
	\centering
	\includegraphics[width=\linewidth]{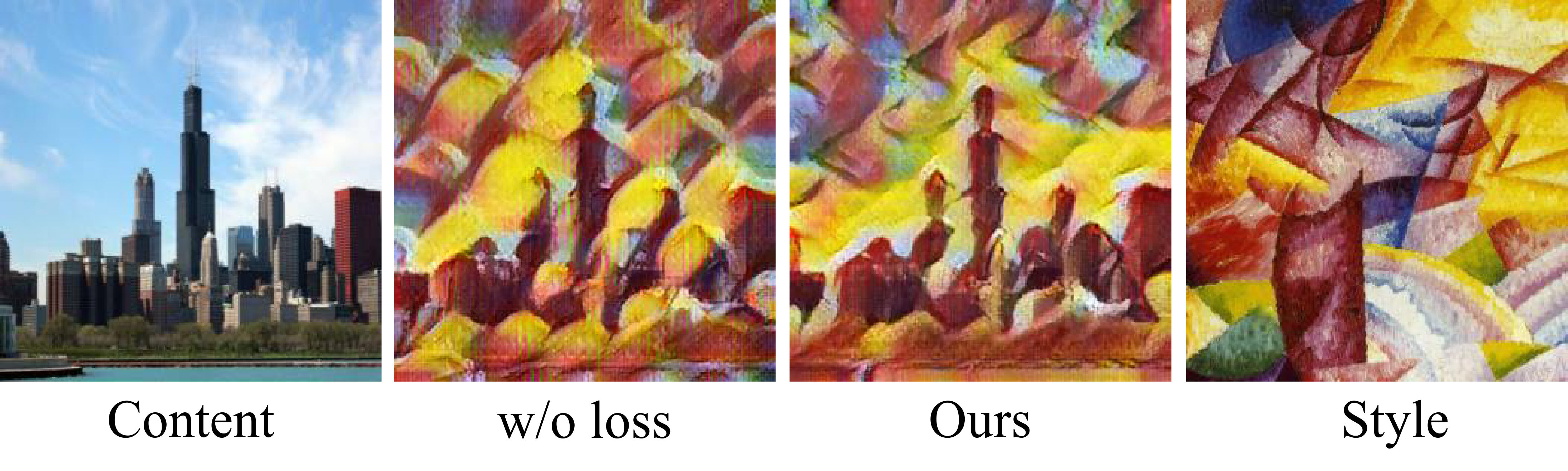}
	\caption{Ablation study on \emph{whitening loss}. With the proposed loss, clearer content contours and better style pattern consistency are achieved.}
	\label{Fig:ab_whiting_loss}
\end{figure}

Assessing style transfer results could be subjective. We thus conduct two quantitative comparisons, which are reported in first 2 rows of Table~\ref{tab:p_loss}. We first compares different methods in terms of perceptual loss. This evaluation metrics contain both content and style terms which have been used in previous approaches \cite{Huang2017ArbitraryST}. It is worth noting that our approach does not minimize perceptual loss directly since it is only one of the three types of losses we use.  Nevertheless, our model achieves the lowest perceptual loss among all feed-forward models, with style loss being the lowest and content loss only slightly higher than AdaIN. This indicates our approach favors fully stylized results over results with high content fidelity.

We then conduct a user study to evaluate the visual preference of the six methods. 30 content images and 30 style images are randomly selected from the test set and 900 stylization results are generated for each method. Then results of the same stylization are randomly chosen for a participant who is asked to vote for the method that achieves the best stylization. Each participant is asked to do 20 rounds of comparison. The stylized results from different methods are exhibited in a random order. Thus we collect 600 votes from 30 subjects. The average preference scores of different methods are reported in Column 4 of Table~\ref{tab:p_loss}, which shows our method obtains the highest score.

Table~\ref{tab:p_loss} also lists the running time of our approach and various state-of-the-art baselines. All results are obtained with a 12G Titan V GPU and averaged over 100 $256 \times 256$ test images.
Generally speaking, existing patch based network approaches are known to be slower than the holistic matching methods. Among all the approaches, Li et al. achieves the fastest stylization with a linear propagation module. 
Our full model equiped with multi-scale strategy slightly increases the computation burden but are still comparable to AdaIN, thus achieving style transfer in real-time.

\paragraph{Ablation Study}
Here we respectively evaluate the impacts of common feature learning, the proposed \emph{whitening loss} on content feature, and the multi-scale usages of our framework. 

Common feature disentanglement during joint analysis plays a key role in our approach. Its importance can be evaluated by removing the Feature Exchange block and disabling the feature exchange loss, which prevents the network to learn exchangeable features.
As shown in Figure~\ref{Fig:teaser_comparison}, for the ablated model without common features, the color distribution and texture patterns in the result image no longer mimic the target style image. Visually, our full model yields a much more favorable result. 
We also compares the perceptual losses over 100 test images for both the baseline model (i.e. our model without common features) and our full model. As reported in Table~\ref{tab:p_loss}, the style loss of our full model is significantly improved over the baseline, demonstrating the effectiveness of common features.

To verify the effect of \emph{whitening operation} functioned on content features, we remove learnable matrices $W_{whi}$ at all scales to see how the performance changes. As shown in Figure~\ref{Fig:ab_whiting_loss}, without the purified operation and \emph{whitening loss}, the baseline model blurs the overall contours with yellow blobs. In constrast, our full model better matches the target style to the content image and preserves the spatial structures \& style pattern consistency, yielding more visually pleasing results. This proves that the proposed operation enables the content features to be more compatible with the target styles.

The multi-scale strategy is evaluated by replacing the full model with an alternative model that only fuses content and style at $relu\_4$ layer while fixing the other parts. The comparison shown in Figure~\ref{Fig:ab_multi_scale} demonstrates that the multi-scale strategy is more successful in capturing the salient style patterns, leading to better stylization results.

\paragraph{Applications}
We demonstrate the flexibility of our model using two applications. All these tasks are completed with the same trained model without any further fine-tuning.

Being able to adjust the degree of stylization is a useful feature. In our model, this can be achieved by blending between the stylized feature map $f_{cs}$ 
and the VGG-based feature $f_c$ before feeding to the decoder, which is:
\begin{equation}
	F = (1 - \alpha) \cdot  f_c + \alpha \cdot  f_{cs}. \nonumber
\end{equation}
By definition, the network outputs the reconstructed image $I_{c \rightarrow c}$ when $\alpha=0$, the fully stylized image $I_{s \rightarrow c}$ when $\alpha=1$, and a smooth transition between the two when $\alpha$ is gradually changed from 0 to 1; see Figure~\ref{Fig:application_tradeOff}.

In Figure~\ref{Fig:application_mask}, we present our model's ability for applying different styles to different image regions. Masks are used to specify the correspondences between different content image regions and the desired styles. Pair-wise exchangeable feature extraction only consider the masked regions when applying a given style, helping to achieve optimal stylization effect for individual regions.

\section{Conclusions}

In this paper, we have presented a novel framework, EFANet, for transferring an arbitrary style to a content image. By analyzing the common style feature from both inputs as a guider for alignment, exchangeable style features are extracted. Better stylization can be achieved for the content image by fusing its purified content feature with the aligned style feature from the style image.  
Experiments show that our method significantly improves the stylization performance over the prior state-of-the-art methods. 

{\small
	\bibliographystyle{aaai}
	\bibliography{EFANet}
}

\begin{figure*}[t]
	\centering
	\includegraphics[width=\linewidth]{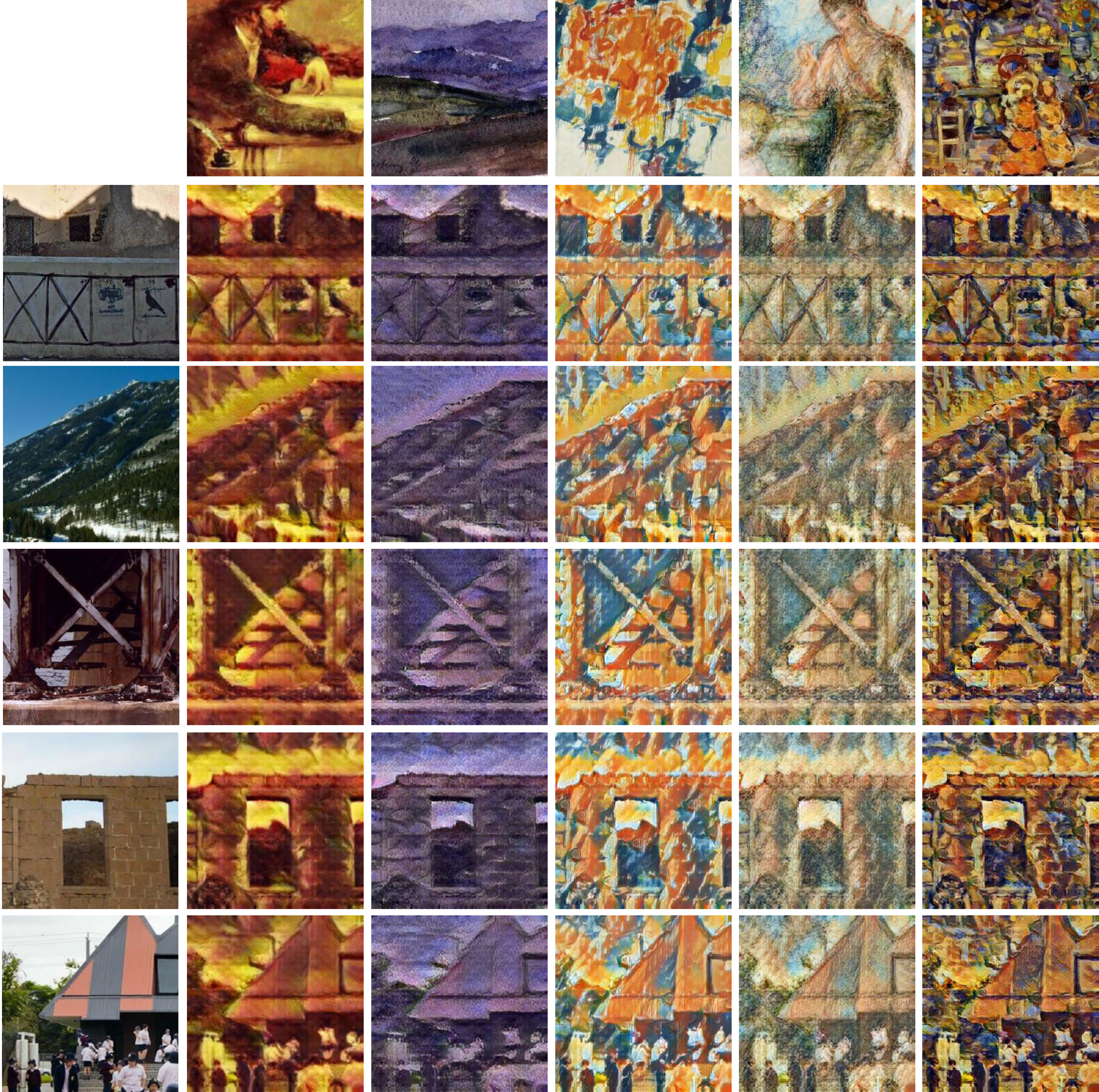}
	\caption{Stylization matrix of transferring different content images to different styles. The first row consists of style images and the content images are listed in the leftmost column.}
	\label{Fig:supp_matrix1}
\end{figure*}

\begin{figure*}[t]
	\centering
	\includegraphics[width=\linewidth]{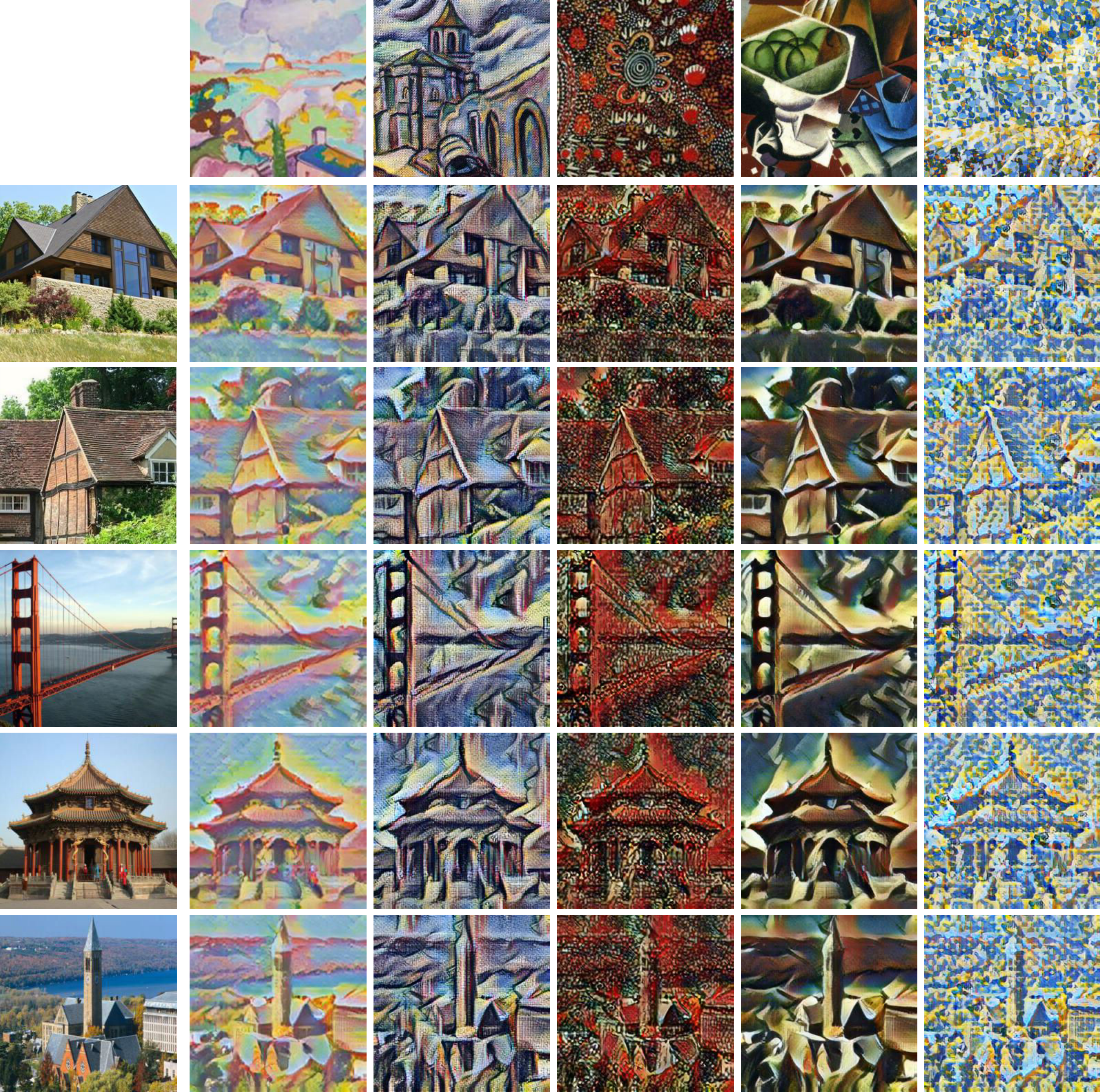}
	\caption{Stylization matrix of transferring different content images to different styles. The first row consists of style images and the content images are listed in the leftmost column.}
	\label{Fig:supp_matrix3}
\end{figure*}

\begin{figure*}[!htb]
	\centering
	{\includegraphics[width=\linewidth]{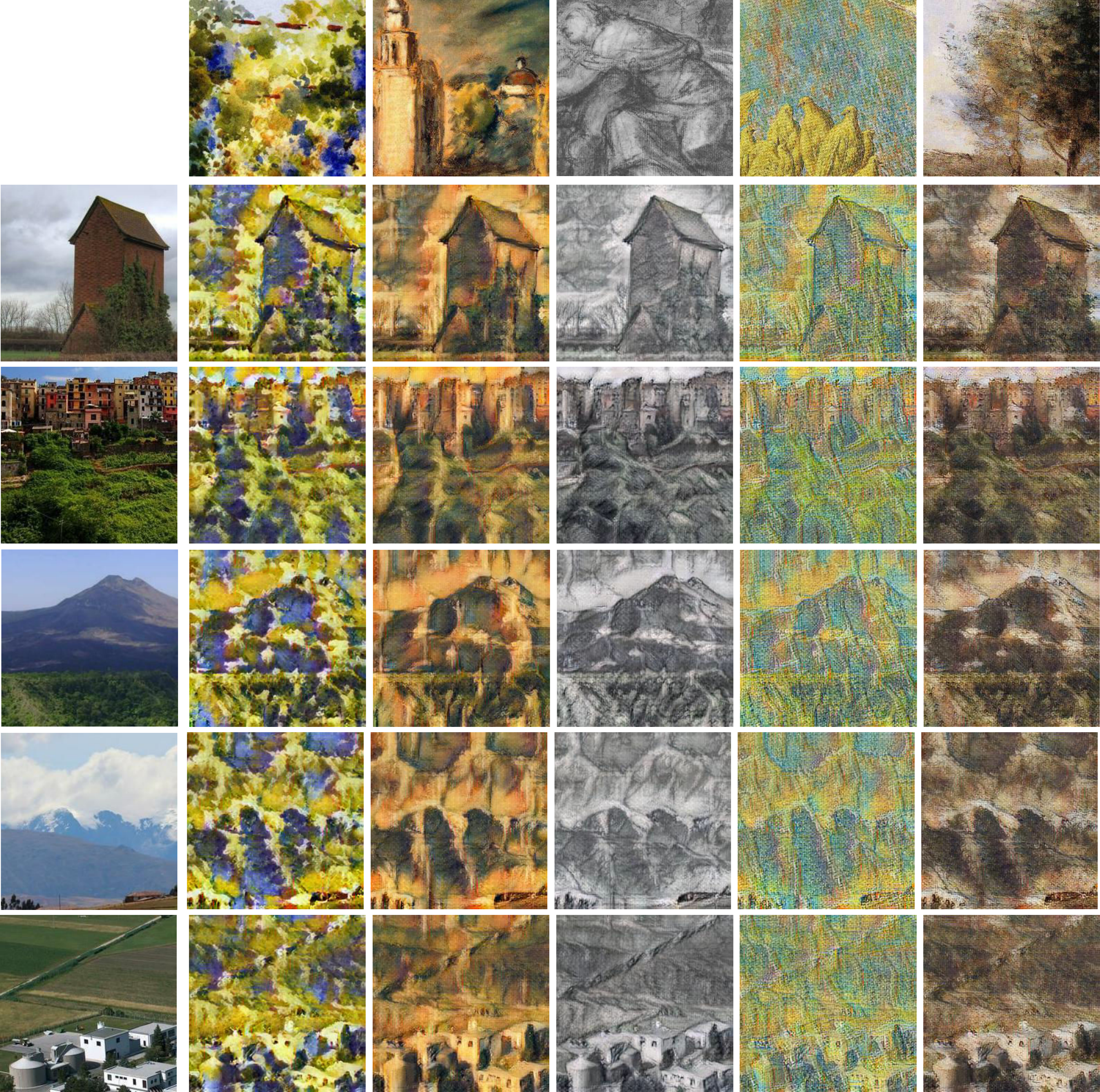}}
	\caption{Stylization matrix of transferring different content images to different styles. The first row consists of style images and the content images are listed in the leftmost column.}
	\label{Fig:supp_matrix2}
\end{figure*}

\end{document}